\documentclass{article}
\usepackage{ijcai23}

\usepackage{times}
\usepackage{soul}
\usepackage{url}
\usepackage[hidelinks]{hyperref}
\usepackage[utf8]{inputenc}
\usepackage[small]{caption}
\usepackage{graphicx}
\usepackage{amsmath}
\usepackage{amsthm}
\usepackage{booktabs}
\usepackage[ruled,vlined,linesnumbered]{algorithm2e}
\usepackage[switch]{lineno}

\usepackage{comment}
\usepackage{subcaption}
\usepackage{xspace}
\usepackage{amsfonts}
\usepackage{mathtools}
\usepackage{xcolor}
\usepackage{enumitem}

\newtheorem{theorem}{Theorem}
\newtheorem{lemma}[theorem]{Lemma}

\title{A Unification Framework for Euclidean and Hyperbolic Graph Neural Networks}

\author{Mehrdad Khatir$^1$, Nurendra Choudhary $^1$,\\{Sutanay Choudhury$^2$, Khushbu Agarwal$^2$, }\\
 $^1$Department of Computer Science, Virginia Tech, Arlington, VA, USA\\
 $^2$Pacific Northwest National Laboratory, Richland, WA, USA\\
 \small{\{khatir, nurendra\}@vt.edu,}\\{\{sutanay.choudhury, khushbu.Agarwal\}@pnnl.gov, reddy@cs.vt.edu}}

\author{
Mehrdad Khatir$^1$\and
Nurendra Choudhary$^1$\and
Sutanay Choudhury$^2$\and
Khushbu Agarwal$^2$\and
Chandan K. Reddy$^{1}$
\affiliations
 $^1$Department of Computer Science, Virginia Tech, Arlington, VA, USA\\
 $^2$Pacific Northwest National Laboratory, Richland, WA, USA\\
\emails
\{khatir, nurendra\}@vt.edu,
\{sutanay.choudhury, khushbu.Agarwal\}@pnnl.gov, 
reddy@cs.vt.edu
}

\begin{document}

\maketitle

\begin{abstract}
Hyperbolic neural networks can effectively capture the inherent hierarchy of graph datasets, and consequently a powerful choice of GNNs. However, they entangle multiple incongruent (gyro-)vector spaces within a layer, which makes them limited in terms of generalization and scalability. 
In this work, we propose the Poincaré disk model as our search space, and apply all approximations on the disk (as if the disk is a tangent space derived from the origin), thus getting rid of all inter-space transformations. Such an approach enables us to propose a hyperbolic normalization layer and to further simplify the entire hyperbolic model to a Euclidean model cascaded with our hyperbolic normalization layer. We applied our proposed nonlinear hyperbolic normalization to the current state-of-the-art homogeneous and multi-relational graph networks. We demonstrate that our model not only leverages the power of Euclidean networks such as interpretability and efficient execution of various model components, but also outperforms both Euclidean and hyperbolic counterparts on various benchmarks. Our code is made publicly available at \url{https://github.com/oom-debugger/ijcai23}.
\end{abstract}

\section{Introduction}

\begin{figure}[tb]
\begin{subfigure}{\columnwidth} 
  \centering
  \includegraphics[width=1\columnwidth]{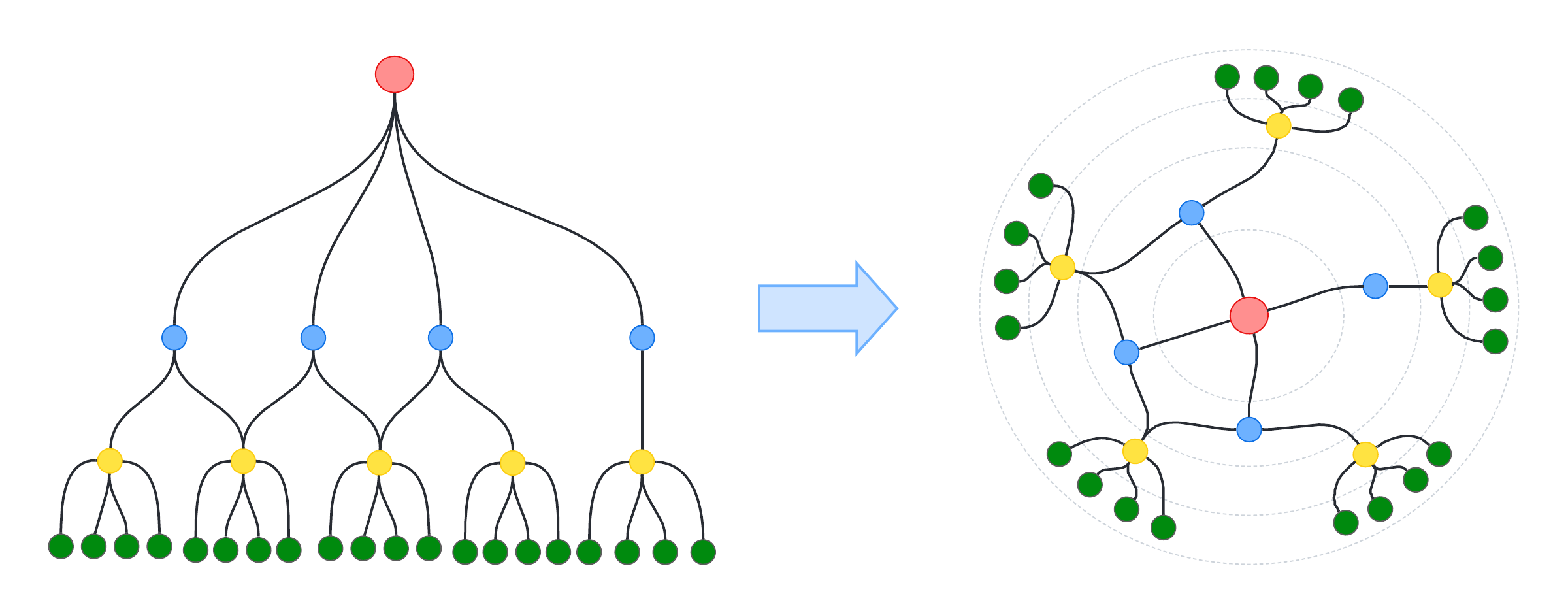}  
  \caption{}
\end{subfigure}
\begin{subfigure}{\columnwidth}
  \centering
  \includegraphics[width=1\columnwidth]{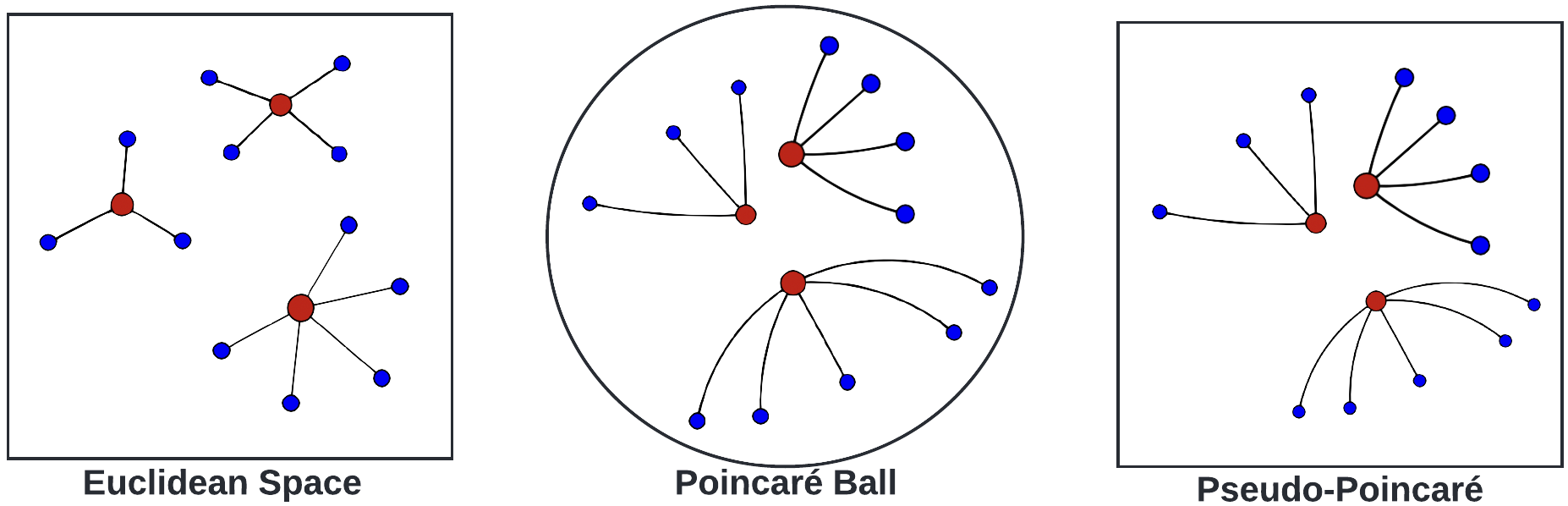} 
  \caption{}
\end{subfigure}
\caption{ (a) Mapping hierarchical data onto a Poincaré ball. Starting with the root as origin (red), the children are continuously embedded with exponentially increasing distances. (b) Aggregation of vectors in Euclidean, Poincaré Ball, and pseudo-Poincaré space (hyperbolic vectors in Euclidean space). {\color{red}Red} points are the aggregation of {\color{blue}blue} points. Pseudo-Poincaré paradigm reformulates hyperbolic operations to capture hierarchy in Euclidean space.}
\label{fig:1}
\end{figure}

Graph data structures are ubiquitous. There have been several advancements in graph processing techniques across several domains such as citation networks \cite{sen2008collective}, medicine \cite{anderson1992infectious}, and e-commerce \cite{choudhary2022anthem}. Recently, non-Euclidean geometries such as hyperbolic spaces have shown significant potential in representation learning and graph prediction \cite{chami2019hyperbolic,choudhary2021self}. Due to their exponential growth in volume with respect to radius \cite{krioukov2010hyperbolic}, hyperbolic spaces are able to embed hierarchical structures with low distortion, i.e., an increase in the number of nodes in a tree with increasing depth is analogous to moving outwards from the origin in a Poincaré ball (since both grow exponentially). This similarity enables the hyperbolic space to project the child nodes exponentially further away from the origin compared to their parents (as shown in Figure \ref{fig:1}.a) which captures hierarchy in the embedding space. There are certain fundamental questions about hyperbolic models which are yet to be studied. For example, hyperbolic models operate on completely new manifolds with their own gyrovector space \cite{2008Ungar} (to be differentiated from the Euclidean vector space). Because they operate in a different manifold, techniques such as dropout or L1/2 regularization do not guarantee similar behavior. Moreover, while existing hyperbolic operations \cite{ganea2018hyperbolic} provide fair support for basic sequential neural networks, a correct translation of more complex Euclidean architectures (such as multi-headed attention networks  \cite{vaswani2017attention}) is non-trivial and counter-intuitive, due to the special properties of Riemannian manifold \cite{shimizu2021hyperbolic}.
Finally, hyperbolic operators are computationally expensive i.e., the number of complex mathematical operations for each primitive operator limits their scalability compared to the Euclidean graph neural networks.

Traditionally, hyperbolic models use a mixture of hyperbolic geodesics, gyro-vector space mathematics (e.g. Poincaré disc model combined with Möbius operations), and tangent space approximations at various points. Instead of switching between multiple frameworks with potentially incongruent operations, 
in this work, we let the  Poincaré disk model be our search space and propose to apply all approximations inplace on the Poincaré disk (i.e., to apply all tangent space approximations on the disk as if the disk is a tangent space derived from the origin). 
This enables us to replace non-scalable gyro-vector/Möbius operations with a Euclidean approximation, and then simplify the whole hyperbolic model as a Euclidean model cascaded with a hyperbolic normalization function. In our approach, we replace all Möbius operations with Euclidean operations, yet we still work in the Riemannian manifold (by using our proposed hyperbolic normalization layer along with using Riemannian optimizer), thus, we call it \textbf{Pseudo-Poincaré} framework
(figure \ref{fig:1}.b).

With our formulations, the trainable parameters lie in the Euclidean space, hence we can re-use the architectures that already exist for Euclidean solution space. This leads to the use of inexpensive operators. It also enables the use of existing well-optimized Euclidean models along with their hyper-parameter settings tuned for specific tasks. Having the aforementioned two capabilities, our framework brings the best of two worlds together. For further evidence, we have applied our formulation to construct new pseudo-Poincaré variants of $GCN$ \cite{kipf2017semi}, $GAT$ \cite{velickovic2018graph}, and multi-relational graph embedding model $MuR$ \cite{balazevic2019multi}, $RGCN$ \cite{schlichtkrull2018modeling} and $GCNII$ \cite{pmlr-v119-chen20v} (we call our variants as $NGCN$, $NGAT$, $NMuR$, $NRGCN$, and $NGCNII$, respectively). We demonstrate that the reformulated variants outperform several state-of-the-art graph networks on standard graph tasks. Furthermore, our experiments indicate that our pseudo-Poincaré variants benefit from using regularization methods used in Euclidean counterparts (unlike the traditional hyperbolic counterparts). We also show the execution speed of our proposed model to be comparable to that of the Euclidean counterparts and hence provide insights for building deeper hyperbolic (pseudo-Poincaré) models.

\section{Related Work}
\label{sec:related}

Research into graph representation learning can be broadly classified into two categories based on the nature of the underlying graph dataset: (i) Homogeneous graphs - where the edges connect the nodes but do not have any underlying information and (ii) Heterogeneous or multi-relational graphs - where the edges contain information regarding the type of connection or a complex relation definition. Early research in this area relied on capturing information from the adjacency matrix through factorization and random walks. In matrix factorization-based approaches \cite{cao2015grarep}, the adjacency matrix $A$ is factorized into low-dimensional dense matrices $L$ such that $\|L^TL-A\|$ is minimized. In random walk-based approaches
\cite{grover2016node2vec,narayanan2017graph2vec}, the nodes' information is aggregated through message passing over its neighbors through random or metapath traversal \cite{fu2020magnn}. More recently, the focus has shifted towards neighborhood aggregation through neural networks, specifically, Graph Neural Networks (GNNs) \cite{scarselli2008graph}. In this line of work, several architectures such as GraphSage \cite{hamilton2017inductive}, GCN \cite{kipf2017semi}, and GAT \cite{velickovic2018graph} have utilized popular deep learning frameworks to aggregate information from the nodes' neighborhood. Concurrently multi-relational graph neural networks have been developed focusing on representation learning in knowledge graphs in which the nodes/entities can have edges/relations of different types. TransE \cite{bordes2013translating}, DistMult \cite{yang2015embedding}, RotatE \cite{sun2018rotate}, MuRE \cite{balazevic2019multi}, RGCN \cite{schlichtkrull2018modeling} are examples of multi-relational GNNs. 

\begin{figure*}[!ht]
  \centering
  \includegraphics[width=0.77\textwidth]{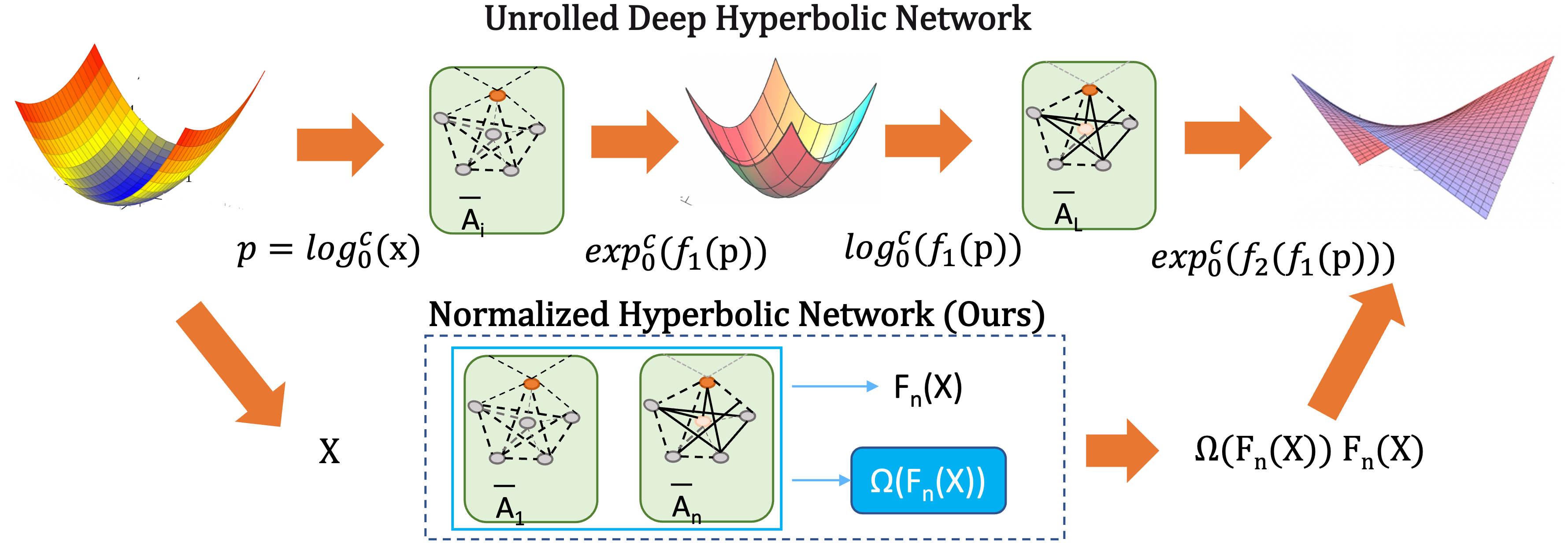}
\caption{ The top row labeled ``Unrolled Deep Hyperbolic Network'' presents the general idea of stacking deep hyperbolic network with $L$ layers.  The input to each layer is passed through a logarithmic map to project onto a Euclidean space, and following the Euclidean layer computation, the output is projected back onto the hyperbolic space. The section \textbf{``Hyperbolic Layers''} 
shows the operator and function chaining technique. Our methodology shown in the lowest layer completely avoids the repeated application of these subspace transformations.}
\label{fig:hyperbolic_normalization_framework}
\end{figure*}

\section{Hyperbolic Networks}
\label{sec:hyperbolic}
Manifold hypothesis suggests that high dimensional data tends to lie in the vicinity of a low dimensional manifold \cite{cayton2005algorithms}\cite{fefferman2016testing}. For such data, the model in fact learns the manifold to better learn the relation between the data points. 
In graph datasets, hierarchical structures are common, which tend to better align in hyperbolic manifolds. Using the aforementioned concept, Ganea, et al. \cite{ganea2018hyperbolic} proposed hyperbolic neural networks. Such a network operates in hyperbolic manifolds, simplifies the search space by omitting the manifold learning process, and thus, results in more efficient models for graph applications. That work led to several architectures for representation learning in both homogeneous (HGCN \cite{chami2019hyperbolic}, HAT \cite{zhang2021hyperbolic}) and heterogeneous (MuRP \cite{balazevic2019multi}, HypE \cite{choudhary2021self}) graph datasets. These architectures primarily utilize the gyrovector space model to formulate the neural network operations.

\textbf{Hyperbolic space} is a Riemannian manifold with constant negative curvature. The coordinates in the hyperbolic space can be represented in several isometric models such as 
Beltrami-Klein Model $\mathbb{K}^n_c$ and Poincaré Ball $\mathbb{B}^n_c$. In this paper, we consider the Poincaré Ball model because it is the most widely used one for hyperbolic networks. 
In a Poincaré ball of curvature $c$, simple operations such as vector addition and scaling are done using Möbius operation (addition $\oplus_c$, {scalar multiplication} $\otimes_c$). More complex operations such as aggregation, convolution, and attention are done in local tangent spaces which require transformation between Euclidean and hyperbolic space. 
For an embedding vector $x\in\mathbb{R}^n$ from point $v$, its corresponding hyperbolic embedding in the Poincaré ball of curvature $c$, $p \in\mathbb{B}^n_c$ is calculated with the exponential map as:
\begin{equation} 
\label{eq1}
p = \exp_v^c(x) = v \oplus_c ({tanh(\frac{\lambda_v||x||}{2})}\frac{x}{||x||}), 
\end{equation}
where $\lambda$ is the conformal factor at point $v$. Conversely, logarithmic map is used to transform a hyperbolic embedding $p \in\mathbb{B}^n_c$ to a Euclidean vector $x\in\mathbb{R}^n$ is formulated as:
\begin{equation} 
\label{eq2-1}
x = \log_v^c(p) = \frac{2}{\lambda_v} tanh^{-1}(||-p\oplus_c v||)\frac{-p\oplus_c v}{||-p\oplus_c v||} .
\end{equation}
Note that the tangent space approximation relies on the assumption of spatial locality (i.e., small vectors starting from point $v$). For transporting a local vector from a point in the Hyperbolic manifold to another point, one can use parallel transport \cite{lee2018introduction}. Parallel transport is specifically used in bias addition in Hyperbolic linear layers \cite{chami2019hyperbolic}. For a detailed discussion on parallel transport, please refer to \cite{lee2018introduction}.

\subsection{Hyperbolic Layers}
\label{app:hyperbolic-layers}
\textbf{Hyperbolic Feed-forward layer.} For the Euclidean function $f:\mathbb{R}^n\rightarrow \mathbb{R}^m$, its equivalent Möbius version $f^{\otimes_c}:\mathbb{B}^n_c\rightarrow \mathbb{B}^m_c$  for the Poincaré ball is defined as $f^{\otimes_c}(p) \coloneqq \exp_0^c\left(f\left(\log_0^c(p)\right)\right) \label{eq:h_ffn}$. Extending this, the hyperbolic equivalent $h^{\mathbb{B}}(p)$ of the Euclidean linear layer $h^{\mathbb{R}}(x) = Wx$ with weight matrix $W \in \mathbb{R}^{n\times m}$ is formulated as:
\begin{align} \label{eq3}
 h^{\mathbb{B}}(p) &= 
 \frac{tanh\left(\frac{|Wp|}{|p|}tanh^{-1}\left(\sqrt{c}|p|\right)\right)}{\sqrt{c}|Wp|} Wp
\end{align}

\noindent \textbf{Hyperbolic Convolution.} For a hyperbolic graph node representation $p_0 \in \mathbb{B}^n_c$ with neighbors $\{p_i\}_{i=1}^k$. As given in \cite{chami2019hyperbolic},  the hyperbolic convolution layer $GCN^{\otimes_c}:\mathbb{B}^n_c \rightarrow \mathbb{B}^m_c$ constitutes of a linear transform, followed by neighborhood aggregation\footnote{The aggregation occurs in the tangent space as concatenation is not well-defined for hyperbolic space.} which is computed as:
\begin{align}
\nonumber h^{\mathbb{B}}(p_i) &= W_i \otimes_c p_i \oplus_c b_i 
\end{align}

\begin{align}
\nonumber h^{\mathbb{B}}_{agg}(p_0) = exp^c_{p_0}\left(\Sigma_{i=0}^kw_{ij}\log^c_{p_0}\left(h^{\mathbb{B}}(p_i)\right)\right) 
\end{align}

\begin{align}
 GCN^{\otimes_c}(p_0) &= exp^c_0\left(\sigma(\log^c_0(h^{\mathbb{B}}_{agg}(p_0))\right) \label{eq:hgcn}
\end{align}
where $w_{ij}= softmax_{i=1}^k(MLP(log^c_0(p_0)||log^c_0(p_i)))$ is an aggregation weight for neighbors.

\vspace{0.08in}
\noindent \textbf{Multi-relational Hyperbolic Representation.} The multi-relational representations for knowledge graph $KG$ are modeled using a scoring function. For a triplet $(h,r,t) \in KG$, where relation $x_r \in \mathbb{R}^{n}$ connects the head entity $x_h \in \mathbb{R}^{n}$ to the tail entity $x_t \in \mathbb{R}^{n}$ and $R \in \mathbb{R}^{n\times n}$ is the diagonal relation matrix, the hyperbolic equivalent $\phi^{\mathbb{B}}(p_h,p_r,p_t)$ of the Euclidean scoring $\phi(x_h,x_r,x_t)=-\|Rx_h-(x_t+x_r)\|^2$ function is computed using the hyperbolic distance $d^\mathbb{B}$ as:
\begin{align}
     d^\mathbb{B}(p_1,p_2) &= 
     \frac{2}{\sqrt{c}}\tanh^{-1}\left(\sqrt{c}\|-p_1\oplus_c p_2\|\right) \label{eq:mrh_dist}
\end{align}

\section{Pseudo-Poincaré Hyperbolic Networks}
\label{sec:bridge}

While hyperbolic networks show promising results, switching back and forth between Möbius gyro-vector operations and Euclidean tangent approximations (at various points) makes them computationally expensive, limits their extensibility, and makes them less predictable when applying L1/2 regularization and dropout. Furthermore, such back and forth between the manifolds (at variable points) comes with an implicit assumption that the embeddings have spatial locality in all dimensions for the models to be performant. 
We argue that approximating on tangent space at a variable point does not necessarily increase the approximation accuracy because the spatial locality of the embeddings is not guaranteed in general. Therefore, fixating tangent space to be at the origin can bring both simplicity to the model and performance in general cases.
Moreover, only tangent space at origin has a one-to-one mapping with all points on the Poincaré (hyperbolic) half-plane model. This enables us to combine the tangent space approximation with Poincaré half-plane model and apply all approximations in the half-plane. This makes the pseudo-Poincaré networks much easier to understand, develop, and optimize. Fixating the tangent space to the point of origin simplifies the exponential maps (eq. \ref{eq1}) as:
\begin{equation}
\label{eq1:simp}
p = \exp_0^c(x) = \frac{tanh(\sqrt{c}|x|)}{\sqrt{c}|x|} x.
\end{equation}
Conversely, the logarithmic map (eq. \ref{eq2-1}) is simplified as:

\begin{equation}
\label{eq2-1:simp}
x = \log_0^c(p) = \frac{tanh^{-1}(\sqrt{c}|p|)}{\sqrt{c}|p|}p.
\end{equation}
Using Eq. (\ref{eq1:simp}) and Eq. (\ref{eq2-1:simp}), we reformulate the hyperbolic feed-forward layer. In this section, we only discuss the re-formulation for Pseudo-Poincaré Feed-forward Layer. The convolution, attention, and multi-relation layer can be similarly reformulated for the Pseudo-Poincaré paradigm.

\vspace{0.05in}

\noindent \textbf{Pseudo-Poincaré Feed-forward Layer.} We state that cascading hyperbolic feed-forward layers (including linear layers) is equivalent to cascading their Euclidean counterparts and then applying a hyperbolic normalization to the result.

\begin{lemma}
\label{lm:simp_exp}
For a point in the tangent space at origin $x\in\mathcal{T}_{0_n}\mathbb{B}^n_c$, the exponential map to the hyperbolic space $\mathbb{B}^n_c$, $\exp_0^c:\mathcal{T}_{0_n}\mathbb{B}^n_c\rightarrow \mathbb{B}^n_c$ $\exp_0^c(x)$, is equivalent to the input scaled by a non-linear scalar function $\omega:\mathbb{R}^n\rightarrow\mathbb{R}^n$, i.e.,
\begin{equation} \label{eq2}
p = \exp_0^c(x) = \omega(x)x;\quad \omega(x) = \frac{tanh(\sqrt{c}|x|)}{\sqrt{c}|x|}
\end{equation}
where $\omega(x)$ is the hyperbolic normalization for exponential maps.  This follows from equation (1).
\end{lemma}

\begin{lemma}
\label{lm:ff} 
Let $f^{\otimes_c}$ be a single hyperbolic feed-forward layer, and $f^{\otimes_c}$ be its euclidean counterpart.
Approximating on tangent space at origin for all stages makes the cascade of $n$ hyperbolic feed-forward layers $F_n^{\otimes_c} = \{f^{\otimes_c}_i\}_{i=1}^n$ to be equivalent to cascading $n$ of their Euclidean counterparts $F_n = \{f_i\}_{i=1}^n$ encapsulated in $\log_0^c$ and $\exp_0^c$ functions, i.e.,
\begin{align}
    F_n(x) &= f_n\left(f_{n-1}\left( \cdots f_1\left(x\right)\right)\right)\\
   \nonumber F_n^{\otimes_c}(p) &= f_n^{\otimes_c}(f^{\otimes_c}_{n-1}( \cdots f_1^{\otimes_c}(p)) = exp_0^c\left(F_n(log^c_0(p))\right)
\end{align}
\end{lemma}

\noindent \textbf{Proposition} Lemma \ref{lm:ff} implies that the hyperbolic transformations $\exp_0^c$ and $\log_0^c$ are only needed at the initial and final stages to capture hierarchy in Euclidean networks.

\begin{lemma}
\label{lm:input-features} For given input features $x=log_0^c(p) \in \mathbb{R}^n$, a given hyperbolic feed forward layer can be rewritten as $f^{\otimes_c}(x) \coloneqq \omega(f(x))f(x)$.
\end{lemma}
From lemmas \ref{lm:ff} and \ref{lm:input-features}, we arrive at the following theorem for cascaded hyperbolic layers.
\begin{theorem}
\label{ff-generalized}
Fixating the tangent space to be always at origin, cascaded hyperbolic feed-forward layers $F_n^{\otimes_c}$ for Euclidean input $x\in \mathbb{R}^n$ can be reformulated as $F_n^{\otimes_c}(x) \coloneqq \Omega(F_n(x))F_n(x)$, where $\Omega(F_n(x))$ is the hyperbolic normalization for a cascaded hyperbolic feed-forward network. Note that  $\omega()$ is the exponential transformation factor at the variable level and  $\Omega()$ is the same at the level of neural networks.
\end{theorem}

\noindent \paragraph{Hyperbolic Normalization.} Theorem \ref{ff-generalized} allows the cascaded hyperbolic layers to be reformulated in the Euclidean space using a hyperbolic normalization ($\Omega$) function. 
Our normalization layer resembles the technique proposed in \cite{salimans2016weight} to normalize the weights, which uses the equation; $\label{w-norm} w = \frac{e^s}{\|v\|}v$, where $v$ is an $n$-dimensional vector used to learn the direction and $s$ is a trainable parameter. However, unlike weight normalization, hyperbolic normalization happens on the features (and not the weights). Furthermore, hyperbolic normalization uses the $tanh$ scaling function instead of $e^s$.
However, both are similar in terms of using the feature vector magnitude as the denominator and hence removing the orthogonality of the dimensions.
The ``unrolled deep hyperbolic network'' 
(shown in Figure \ref{fig:hyperbolic_normalization_framework})
presents the general idea of stacking multiple hyperbolic layers together. The input into each layer is passed through a logarithmic map to project into Euclidean space and  its output is projected back onto the hyperbolic space after applying the computations.  

\noindent \paragraph{Batch-Norm, Layer-Norm vs. Hyperbolic Norm.} 
Batch-Norm tries to reduce the covariant shift and noise \cite{ioffe2015batch} which means the result of the batch-norm is in Euclidean space. On the other hand, Layer-Norm, and Hyperbolic Norm take vectors to another manifold for better latent space representation \cite{ba2016layer}. Thus, we recommend using Riemannian optimizers when applying our method at least for shallow GNNs (we provided empirical results in the next section).
Note that, in order to apply the batch normalization in hyperbolic space, one needs to change its mathematical operations to use Frechet mean \cite{peng2021hyperbolic}. The detailed analysis of the effect of hyperbolic batch normalization is outside the scope of this work.

\noindent \paragraph{Optimization.} 
The scaling function in Eq. (\ref{eq2}) uses the vector norm in the denominator and hence, each dimension depends on the value of other dimensions (i.e., non-orthogonal). Due to this, optimizers that are developed based on the orthogonality assumptions of the dimensions such as ADAM are not expected to produce the most optimal weights. Thus, we choose Riemannian optimizers such as Riemannian ADAM \cite{becigneul2018riemannian} for our model training. For traditional hyperbolic models, regularization techniques such as L2-regularization do not work as expected since the weights are in hyperbolic space.  However, the trainable parameters in our formulation are in the Euclidean space. Hence, we can additionally apply regularization techniques that are applied to Euclidean networks. 

Algorithm \ref{alg:metapath} provides the generalized algorithm for our proposed hyperbolic model.
\begin{algorithm}[htbp]
	\SetAlgoLined
	\KwIn{Input Euclidean model $F_L$, sample node $x \in \mathbb{R}^n$}
	\KwOut{Output of Normalized hyperbolic model $F_L^{\otimes_c}(x)$;}
	$f_{0}^{\otimes_c}(x) = x;$\\
	\For{layer $l:1\rightarrow L$}
	{  
	    $x \leftarrow f^{\otimes_c}_{l-1}(x);$\\
	    {\# Layer $l$ of the Euclidean model is $f_l(x)$}\\
	    $\omega(f_l(x)) = \frac{tanh(\sqrt{c}|f_l(x)|)}{\sqrt{c}|f_l(x)|}$; using Eq. (\ref{eq2})
	    $f^{\otimes_c}_l(x) = \omega(f_l(x))f_l(x)$; using Lemma \ref{lm:input-features}\\
	}
	$F_n^{\otimes_c}(x) = f^{\otimes_c}_L(x);$\\
	\Return{$F_L^{\otimes_c}(x)$}
	\caption{Normalized hyperbolic model.}
	\label{alg:metapath}
\end{algorithm}

\subsection{Pseudo-Poincaré Layers}
In this section, we dive into the details of a few Pseudo-Poincaré Layers.

\noindent \textbf{Pseudo-Poincaré Graph Convolution.} The individual components of the Hyperbolic Graph Convolution, as defined in Eq. (\ref{eq:hgcn}), can be reformulated as a cascaded hyperbolic feed-forward layer. Thus, we approximate both the aggregation weights and the aggregation function in $\mathcal{T}_0\mathbb{B}^n_c$ as:
\begin{align}
h^{\mathbb{B}}_{agg}(p_i) &= exp^c_0\left(\Sigma_{i=0}^kw_{ij}\log^c_0\left(h^{\mathbb{B}}(p_i)\right)\right) \label{eq:agg}
\end{align}
With this approximation, we apply Theorem \ref{ff-generalized} to reformulate reformulate the $GCN^{\otimes_c}:\mathbb{B}^n_c \rightarrow \mathbb{B}^m_c$ layers as Normalized GCN $NGCN^{\otimes_c}:\mathbb{R}^n \rightarrow \mathbb{R}^m$ over Euclidean inputs $x_0 \in \mathbb{R}^n$ with neighbors $\{x_i\}_{i=1}^k$.  This is computed as $NGCN^{\otimes_c} \coloneqq \Omega(GCN(x_0))GCN(x_0)$, where $GCN(x_0)$ is the Euclidean Graph Convolution layer \cite{kipf2017semi}.

Note that the cascaded feed-forward functions of HGCN \cite{chami2019hyperbolic} operate on different tangent spaces, i.e., aggregation weights are calculated at the linear tangent space at origin $\mathcal{T}_0\mathbb{B}^n_c$ whereas the aggregation function as given in Eq. (\ref{eq:agg}) is driven from a linear tangent space at the root node $\mathcal{T}_{p_0}\mathbb{B}^n_c$. An argument to use different tangent spaces for different function is based on the fact that Euclidean approximation is most accurate when the tangent space is close to the true resulting point. For aggregation, using tangent space at $p_{0}$ requires an implicit assumption that all aggregated hyperbolic points preserve a form of spatial locality, which implies similarity and not exactly the notion of hierarchy. However, we can still argue that hierarchical structures have a loose form of spatial locality, and hence using $log^{c}_0(p)$ will add some noise to the embedding compared to when $log^{c}_{p_0}(p)$ is used. 

\vspace{0.05in}
\noindent
\textbf{Pseudo-Poincaré Graph Attention Layer.}\label{sec:model} Here, our goal is to project the Euclidean GAT layer onto the hyperbolic space by applying the hyperbolic normalization (e.g., Eq. (\ref{eq2})), then, investigate its approximation in relation to the existing hyperbolic graph attention layers as well as with a true hyperbolic graph attention mechanism. 

Hyperbolic attention is formulated as a two-step process (as defined in \cite{gulcehre2018hyperbolic}); (1) matching, which computes attention weights, and (2) the aggregation, which takes a weighted average of the values (attention coefficients as weights). 
Hence, the attention layer can be seen as two cascaded layers, in which the first calculates the attention weights, while the second layer aggregates the weighted vectors. 
Existing hyperbolic graph attention method \cite{zhang2021hyperbolic} $GAT^{\otimes_c}:\mathbb{B}^n_c \rightarrow \mathbb{B}^m_c$ views the hyperbolic aggregation as a hyperbolic feed forward layer with a Euclidean weighted average as its function. Similarly $GAT:\mathbb{R}^n\rightarrow\mathbb{R}^m$ \cite{velickovic2018graph} also uses weighted-average for the aggregation. 

Since we are using Euclidean weighted average over cascaded feed-forward layers, we can apply Theorem \ref{ff-generalized} to hyperbolic GAT and approximate it in the hyperbolic space with a hyperbolic normalization function and Euclidean GAT as $NGAT^{\otimes_c}(x) \coloneqq \Omega(GAT(x))GAT(x)$.

\vspace{0.05in}
\noindent \textbf{Pseudo-Poincaré Multi-relational Representation.} For Euclidean inputs $x_1,x_2 \in \mathbb{R}^n$ and L1-norm $d^\mathbb{R}(x_1,x_2)$, the hyperbolic formulation of the multi-relational hyperbolic distance $d^\mathbb{B}(x,y)$, given in Eq. (\ref{eq:mrh_dist}) can alternatively be reformulated as $d^\mathbb{B}(x_1,x_2) = \omega(x)d^\mathbb{R}(x_1,x_2)$. 
 Similarly, for Euclidean triplet $(x_h,x_r,x_t) \in \mathbb{R}^n$, the scoring function $\phi^{\mathbb{B}}(p_h, p_r, p_t)$ can also be approximated to $NMuR^{\mathbb{B}}(x_h, x_r, x_t)$ as:
 \begin{align}
 \phi^{\mathbb{R}}(x_h, x_r, x_t) &= -d^{\mathbb{R}}\left(Rx_h,x_t+x_r\right)^2 \label{eq:real_score}\\
 \scriptstyle NMuR^{\mathbb{B}}(x_h, x_r, x_t) &= \Omega(\phi^{\mathbb{R}}(x_h, x_r, x_t))\phi^{\mathbb{R}}(x_h, x_r, x_t) \label{eq:new_score}
 \end{align}

\section{Experimental Results}\label{sec:experiments}
In this section, we aim to answer following research questions (RQs) through our experiments:
{\renewcommand\labelitemi{}
\begin{itemize}[noitemsep,leftmargin=*]
    \item \textbf{RQ1:} How do the pseudo-Poincaré variants compare to their Euclidean and hyperbolic counterparts on on node classification and link prediction?
    \item \textbf{RQ2:} How does the pseudo-Poincaré variant, NMuR, compare to its Euclidean and hyperbolic counterparts on reasoning over knowledge graphs?
    \item \textbf{RQ3:} How do pseudo-Poincaré variants react to the choice of optimizers, and what are the effects on their execution time? How does the hyperbolic normalization compare to its counterparts?
    \item \textbf{RQ4:} How does pseudo-Poincaré variant perform when used in deep GNN archictures e.g. GCNII? 
    \item \textbf{RQ5:} What is the difference in the representations obtained through hyperbolic and pseudo-Poincaré projections?
\end{itemize}}

\subsection{Experimental Setup}
For comparison of hyperbolic normalization with other methods, we conduct our experiments on the following tasks: (i) graph prediction (node classification and link prediction) and (ii) reasoning over knowledge graphs, which are described below:

\begin{table*}[tbp]
\centering
    \small
    \begin{tabular}{ll|ccc|ccc}
    \hline
    \textbf{Repr.}&\textbf{Datasets}&\multicolumn{3}{c|}{\textbf{Node Classification (Accuracy in \%)}}&\multicolumn{3}{c}{\textbf{Link Prediction (ROC in \%)}}\\
    \textbf{Space}& \textbf{Models}&\textbf{CORA}&\textbf{Pubmed}&\textbf{Citeseer}&\textbf{Cora}&\textbf{Pubmed}&\textbf{Citeseer}\\\hline
\textbf{Euclidean}&\textbf{GCN}&80.1$\pm$0.3&78.5$\pm$0.3&71.4$\pm$0.3&90.2$\pm$0.2&92.6$\pm$0.2&91.3$\pm$0.2\\
&\textbf{GAT}&\underline{82.7$\pm$0.2}&79.0$\pm$0.3&71.6$\pm$0.3&89.6$\pm$0.2&92.4$\pm$0.2&\textbf{93.6$\pm$0.2}\\
\hline
\textbf{Hyperbolic}&\textbf{HGCN}&77.9$\pm$0.3&78.9$\pm$0.3&69.6$\pm$0.3&\textbf{91.4$\pm$0.2}&\textbf{95.0$\pm$0.1}&\underline{92.8$\pm$0.3}\\
&\textbf{HGAT}&79.6$\pm$0.3&\textbf{79.2$\pm$0.3}&68.1$\pm$0.3&90.8$\pm$0.2&93.9$\pm$0.2&92.2$\pm$0.2\\
\hline
\textbf{Pseudo-}&\textbf{NGCN}&82.4$\pm$0.2&78.8$\pm$0.3&\underline{71.9$\pm$0.3}&\underline{91.3$\pm$0.2}&\underline{94.7$\pm$0.1}&\underline{92.8$\pm$0.2}\\
\textbf{Poincaré}&\textbf{NGAT}&\textbf{83.1$\pm$0.2}&\underline{79.1$\pm$0.3}&\textbf{73.8$\pm$0.3}&90.5$\pm$0.2&93.9$\pm$0.2&\underline{92.8$\pm$0.2}\\\hline
    \end{tabular}
\caption{ Performance comparison results between hyperbolic normalization (ours) and the baseline methods on the graph prediction tasks of node classification and link prediction. The columns present the evaluation metrics, which are Accuracy (for node classification) and Area under ROC (ROC) (for link prediction), along with their corresponding 95\% confidence intervals. {The cells with the best performance are highlighted in bold and the second-best performance are underlined.}} 
\label{tab:overal-perf}
\end{table*}


\vspace{0.05in}
\noindent
\textbf{Graph Prediction.} Given a graph $\mathcal{G}=(V\times E)$, where $v_i \in \mathbb{R}^n, \{v_i\}_{i=1}^{|V|} \in V$ is the set of all nodes and $E \in \{0,1\}^{|V|\times|V|}$ is the Boolean adjacency matrix, where $|V|$ is the number of nodes in the graph and $e_{ij}=1$, if link exists between node $i$ and $j$, and $e_{ij}=0$, otherwise. Each node $v_i \in V$ is also labeled with a class $y_i$.  
In the task of \textbf{node classification}, the primary aim is to estimate a predictor model $P_\theta$ with parameters $\theta$ such that for an input node $v_i$; $P_\theta(v_i|E) = y_i$.
In the task of \textbf{link prediction}, the primary aim is to estimate a predictor model $P_\theta$ with parameters $\theta$ such that for an input node-pair $v_i, v_j$ and an incomplete adjacency matrix $\hat{E}$; $P_\theta(v_i,v_j|\hat{E})= e_{ij}$, where $e_{ij} \in E$ is an element in the complete adjacency matrix.


\vspace{0.05in}
\noindent
\textbf{Reasoning over Knowledge Graphs.} {Let us say that a set of triplets constitutes a knowledge graph $\mathcal{KG} = \{(h_i,r_i,t_i)\}_{i=1}^{|\mathcal{KG}|}$, where $h_i \in \mathbb{R}^n$ and $t_i \in \mathbb{R}^n$ are the head and tail entity, respectively connected by a relation $r_i \in \mathbb{R}^n$. In the task of \textbf{reasoning} \cite{balazevic2019multi}, our goal is to learn representations for the entities and relations using a scoring function that minimizes the distance between heads and tails using the relation as a transformation. The scoring function for Euclidean $MuRE$, hyperbolic $MuRP$ and pseudo-Poincaré $NMuR$ are presented in Eq. (\ref{eq:real_score}),
and Eq. (\ref{eq:new_score}), respectively.}

\subsection{Datasets and Baselines}
\textbf{Datasets.} For comparing our model with the baselines, we chose the following standard benchmark datasets; 
\begin{itemize}
    \item \textbf{Cora} \cite{rossi2015the} contains 2708 publications with paper and author information connected by citation links and classified into 7 classes based on their research areas.
    \item \textbf{Pubmed} \cite{sen2008collective} contains medical publications pertaining to diabetes labeled into 3 classes. The dataset is collected from the Pubmed database.
    \item \textbf{Citeseer} \cite{sen2008collective} contains 3312 annotated scientific publications connected by citation links and classified into 6 classes of research areas.
    \item \textbf{WN18RR} \cite{dettmers2018convolutional} is a subset of the hierarchical WordNet relational graph that connects words with different types of semantic relations. WN18RR consists of 40,943 entities connected by 11 different semantic relations. 
    \item \textbf{FB15k-237} \cite{bordes2013translating} \cite{toutanova2015representing} contains triple relations of the knowledge graph and textual mentions from the Freebase entity pairs, with all simple invertible relations are removed for better comparison. The dataset contains 14,541 entities connected by 237 relations.
    \item Protein-Protein Interaction (\textbf{PPI}) network dataset \cite{hamilton2017inductive} contains 12M+ triples with 15 relations (preserving the original ratio of the relations in the dataset, we randomly sampled the dataset, and trained on the sampled subset). 
\end{itemize}

We use Cora, Pubmed, and Citeseer for our experiments on homogeneous graph prediction, i.e., node classification and link prediction. For comparing the multi-relational networks on the task of reasoning, we utilize the FB15k-237 and WN18RR datasets. For a fair comparison of evaluation metrics, we use the same training, validation, and test splits as used in the previous methods, i.e., the splits for Cora, Pubmed, and Citeseer are as given in \cite{chami2019hyperbolic} and for FB15k-237 and WN18RR, the splits are in accordance with \cite{balazevic2019multi}.

\vspace{0.05in}
\noindent \textbf{Baselines.} For comparing the performance of our Pseudo-Poincaré models, we utilize the Euclidean and hyperbolic variants of the architecture as our baselines; 

\begin{itemize}
    \item  \textbf{Graph Convolution (GCN)} \cite{kipf2017semi} aggregates the message of a node's k-hop neighborhood using a convolution filter to compute the neighbor's significance to the root.
    \item \textbf{Graph Attention (GAT)} \cite{velickovic2018graph} aggregates the messages from the neighborhood using learnable attention weights. 
    \item \textbf{Hyperbolic Graph Convolution (HGCN)} \cite{chami2019hyperbolic} is a hyperbolic formulation of the GCN network that utilizes the hyperbolic space and consequently, Möbius operations to aggregate hierarchical features from the neighborhood,
    \item \textbf{Hyperbolic Graph Attention (HGAT)} \cite{yang2021hgat} is a hyperbolic formulation of the GAT networks to capture hierarchical features in the attention weights.
    
    \item \textbf{Multi-relational Embeddings (MuRE)} \cite{balazevic2019multi} transforms the head entity by a relation matrix and then learns representation by minimizing L1-norm between head and tail entities translated by the relation embedding. 
    \item \textbf{Multi-relational Poincaré (MuRP)} \cite{balazevic2019multi} is a hyperbolic equivalent of MuRE that transforms the head entity by a relation matrix in the Poincaré ball and then learns representation by minimizing the hyperbolic distance between the head and tail entities translated by the relation embedding.
    \item \textbf{Relational Graph Convolution Network (RGCN)} \cite{schlichtkrull2018modeling} is graph convolutional networks applied in heterogeneous graph.
    \item \textbf{Deep Graph Convolutional Networks (GCNII)} \cite{pmlr-v119-chen20v} is an extension of a Graph Convolution Networks with two new techniques, initial residual and identify mapping, to be used in deep GNN settings.
\end{itemize}

\begin{table*}[tbp]
\centering
    \resizebox{\textwidth}{!}{ 
    \begin{tabular}{p{.8em}p{2.8em}|ccc|ccc|ccc}
    \hline
&\textbf{}&\multicolumn{3}{c|}{\textbf{WN18RR}}&\multicolumn{3}{c|}{\textbf{FB15K-237}}&\multicolumn{3}{c}{\textbf{PPI}}\\\hline
\textbf{Dim}&\textbf{Models}&\textbf{MRR}&\textbf{H@10}&\textbf{H@3}&\textbf{MRR}&\textbf{H@10}&\textbf{H@3}&\textbf{MRR}&\textbf{H@10}&\textbf{H@3}\\\hline
\textbf{40}&\textbf{MuRE}&40.9$\pm$0.3&49.7$\pm$0.3&44.5$\pm$0.3
&29.0$\pm$0.3&46.2$\pm$0.3&31.9$\pm$0.3& 2.81$\pm$0.3&9.24$\pm$0.3&4.11$\pm$0.3\\
&\textbf{MuRP}&42.5$\pm$0.3&52.2$\pm$0.3&45.9$\pm$0.3
&29.8$\pm$0.3&47.4$\pm$0.3&32.8$\pm$0.3& 5.55$\pm$0.3&10.21$\pm$0.3&4.43$\pm$0.3
\\
&\textbf{NMuR}&\textbf{43.6$\pm$0.3}&\textbf{57.5$\pm$0.3}&\textbf{47.4$\pm$0.3}
&\textbf{31.5$\pm$0.3}&\textbf{49.7$\pm$0.3}&\textbf{34.7$\pm$0.3}&\textbf{8.21$\pm$0.3}&\textbf{14.3$\pm$0.3}&\textbf{11.7$\pm$0.3}
\\\hline
\textbf{200}&\textbf{MuRE}&44$\pm$0.3&51.3$\pm$0.3&45.5$\pm$0.3
&31.4$\pm$0.3&49.5$\pm$0.3&34.5$\pm$0.3&10.68$\pm$0.3&14.55$\pm$0.3&11.90$\pm$0.3
\\
&\textbf{MuRP}&44.6$\pm$0.3&52.4$\pm$0.3&46.2$\pm$0.3
&31.5$\pm$0.3&49.8$\pm$0.3&34.8$\pm$0.3& 8.23$\pm$0.3&15.09$\pm$0.3&9.98$\pm$0.3
\\
&\textbf{NMuR}&\textbf{44.7$\pm$0.3}&\textbf{57.9$\pm$0.3}&\textbf{48.1$\pm$0.3}
&\textbf{32.2$\pm$0.3}&\textbf{50.6$\pm$0.3}&\textbf{35.5$\pm$0.3}& \textbf{12.68$\pm$0.3}&\textbf{15.07$\pm$0.3}&\textbf{13.82$\pm$0.3}
\\\hline
\end{tabular}}
\caption{ Performance comparison results between Pseudo-Poincaré (ours) and the baseline methods on the task of multi-relational graph reasoning. The columns present the evaluation metrics of Hits@K (H@K) (\%) and Mean Reciprocal Rank (MRR) (\%) along with their corresponding 95\% confidence intervals. The best results are highlighted in bold.}
\label{tab:multi-relation}
\end{table*}

\begin{table}[htbp!]
\centering
\small
\centering
    \begin{tabular}{l|cccc}
    \hline
    \textbf{Dataset} &\textbf{AIFB} & \textbf{AM} & \textbf{BGS}& \textbf{MUTAG}\\\hline
    \textbf{RGCN} & \textbf{97.22} & 89.39 & 82.76 & 67.65  \\
    \textbf{NRGCN}  & \textbf{97.22} &	\textbf{89.9} & \textbf{89.66} & \textbf{73.53} \\ \hline
    \textbf{E-RGCN}  & 88.89 & 90.4	& 82.76	& 75.00 \\
    \textbf{NE-RGCN}  & \textbf{94.44} &	\textbf{90.91} &	\textbf{86.21} &	\textbf{76.47}\\
  \hline
  \end{tabular}  
\caption{ Node-Classification results for RGCN model on AIFB, AM, BGS, and MUTAG datasets. Note that we used the same implementation and dataset in \protect\cite{Thanapalasingam21} to show that our method can be applied incrementally and effectively to the existing code-bases.} 
\label{tab:RGCN-1}
\end{table}

\begin{table}[htbp!]
\small
\centering
\centering
    \begin{tabular}{l|cccc}
    \hline
    \textbf{} & MPR & Hits@1 & Hits3 & Hits@10 \\ \hline
    \textbf{Dataset} & \multicolumn{4}{c}{FB-Toy} \\  \hline
    \textbf{RGCN} & 47.0 &	30.9	& 55.3 & 83.2 \\
    \textbf{N-RGCN}  & \textbf{51.8} & \textbf{35.2} & \textbf{60.9} & \textbf{86.5}  \\ \hline
    \textbf{Dataset} &  \multicolumn{4}{c}{WN18} \\  \hline
    \textbf{RGCN} &  65.9 & 43.2 & 89.0 & 93.7  \\
    \textbf{N-RGCN} & \textbf{75.2} & \textbf{59.8} & \textbf{90.2} & \textbf{93.9} \\ \hline
    \end{tabular} 
\caption{Link-Prediction results for RGCN model on WN18 and FB-Toy datasets. Note that we used the same implementation and dataset in \protect\cite{Thanapalasingam21} to show that our method can be applied incrementally and effectively to the existing code-bases.}
\label{tab:RGCN-2}
\end{table}

\subsection{Implementation Details}
\label{app:implementation-details}
Our models are implemented on the Pytorch framework \cite{paszke2019pytorch} and the baselines are adopted from \cite{chami2019hyperbolic} for GAT, GCN, and HGCN, and from \cite{balazevic2019multi} for MuRE and MuRP. Due to the unavailability of a public implementation, we use our own implementation of HGAT based on \cite{yang2021hgat}. The implementations have been thoroughly tuned with different hyper-parameter settings for the best performance. 

For GAT baseline, we used 64 dimensions with 4 and 8 attention heads (we found 4 heads perform better and hence we only present that result in our tables), with dropout rate of 0.6 and L2 regularization of $\lambda = 0.0005$ (for Pubmed and Cora), and $\lambda = 0.001$ (for CiteSeer) to be inline with the hyper-parameters reported in \cite{velickovic2018graph}. We used same number of dimensions and hyper-parameters for our NGAT model. For GCN baseline, \cite{velickovic2018graph} used 64 dimensions. we used the same number of dimensions and hyperparameter setting as GAT baseline.
It should be noted that, we observed poor performance when we applied regularization techniques (that GAT baseline used) to the hyperbolic models. For HGCN, the original paper used 16 hidden dimension, however, we tried 64, 32, and 16 for the hidden dimensions, we found 64 dimensions produces a better results. 
For the curvature choice, we used different curvatures for different tasks. For the task of node classification, we used $c=0.3$ (for Cora), and $c=1$ (for Pubmed and Citeseer) as they produced the best results, while for the link prediction task, $c=1.5$ produced the best results. at the end, we empirically found that scaling the output of the hyperbolic normalization layer by the factor of 5 produces the best results when $c=0.3-0.5$ (and for $c=1.5$, we set the scaling factor to 3).

\subsection{RQ1: Performance on Graph Prediction}
To evaluate the performance of hyperbolic normalization with respect to Euclidean and hyperbolic alternatives, we compare the $NGCN$ and $NGAT$ models against the baselines on the tasks of node classification and link prediction. We utilize the standard evaluation metrics of accuracy and ROC-AUC for the tasks of node classification and link prediction, respectively (Table \ref{tab:overal-perf}).

From the results, in all cases (except link-prediction on Citeseer), pseudo-Poincaré variants outperform the Euclidean model. Pseudo-Poincaré variants perform better than their hyperbolic counterparts in node classification, but marginally lower in link-prediction. This is inline with our expectations as embeddings of adjacent points in link-prediction on homogeneous graphs tend to have spatial locality. However, spatial locality is less likely for node-classification (as far away nodes still can belong to the same class).
It is also worth noting that  Pseudo-Poincaré outperforms Euclidean in all node-classification datasets regardless of the hyperbolicity of the dataset (e.g., Cora has low hyperbolicity \cite{chami2019hyperbolic} which were previously suggested as a reason that inferior Hyperbolic networks over Euclidean counterparts). This observation suggests that Pseudo-Poincaré is more general purpose framework (i.e., it is performant in a wider variety of tasks/datasets compared to pure-Hyperbolic model).
It should be noted that pseudo-Poincaré variants are consistently at least the second-best performing methods, closely following the top performer which makes it appealing given the low-complexity, and speedup of the model (Table \ref{tab:speed-up}).

\subsection{RQ2: Multi-relational Reasoning}
To compare the performance of hyperbolic normalization against the baselines for multi-relational graphs, we compare the $NMuR$ model against $MuRE$ and $MuRP$ on the task of reasoning. We consider the standard evaluation metrics of Hits@K and Mean Reciprocal Rank (MRR) to compare our methods on the reasoning task. Let us say that the set of results for a head-relation query $(h,r)$ is denoted by $R_{(h,k)}$, then the metrics are computed by $MRR = \frac{1}{n}\sum_{i=1}^n \frac{1}{rank(i)}$ and $Hits@K = \frac{1}{K} \sum_{k=1}^K e_k$, where $e_k= 1$, if $e_k \in R_{(h,r)}$ and $0$, otherwise. Here $rank(i)$ is the rank of the $i^{th}$ retrieved sample in the ground truth. To study the effect of dimensionality, we evaluate the models with two values of the embedding dimensions: $n=40$ and $n=200$. The results from our comparison are given in Table \ref{tab:multi-relation}.

From the results, note that $NMuR$ outperforms the hyperbolic model $MuRP$, on an average, by $\approx 5\%$ with 40 dimensions and $\approx 3\%$ with 200 dimensions across various metrics and datasets. This shows that $NMuR$ is able to effectively learn better representations at lower number of dimensions. Also, note that the difference in performance by increasing the number of dimensions is $<1\%$ for most cases, implying that NMuR is already able to capture the hierarchical space of the knowledge graphs at lower dimensions. This should limit the memory needed for reasoning without a significant loss in performance.

\subsubsection*{Applying Pseudo-Poincare to RGCN}
In order to further show (1) the effectiveness our method, as well as (2) how easy it is to apply our method to an existing model (i.e. adoptability), we took the pytorch models implemented by \cite{Thanapalasingam21}, and added our hyperbolic normalizatoin layer as the last layer of the model. We also trained the model using Riemannian Adam. Table \ref{tab:RGCN-1} shows the comparative results for the Node Classification tasks on AIFB, AM, BGS, and MUTAG datasets (further details about the datasets can be found in \cite{Thanapalasingam21}). We used curvature $c=1.5$, and scaling factor $s=3$ for N-RGCN and $c=0.5$, and  $s=5$ for NE-RGCN (except for MUTAG NE-RGCN that we used $c=1.5$, and $s=3$). We trained the models 10 time and picked the best checkpoint for both Euclidean and our variant. We observed significant improvements in across all metrics when we applied our Hyperbolic normalization method. The fact the applying our method to the existing models is almost seemless increases the ability to explore/adopt hyperbolic models for variety of tasks/models/applications.

Table \ref{tab:RGCN-2} shows the results of Link Prediction over WN18 and FB-Toy datasets (we stuck to the same implementation and the datasets to have apple-to-apple comparison). We only trained RGCN (the pytorch implementation they had in \ref{tab:RGCN-2}). We used curvature $c=1.5$, and scaling factor $s=3$ for FB-Toy, and $c=0.5$, and  $s=5$ for WN18 dataset. We trained the model on WN18 for 7k epochs, and on FB-toy for 12K (to replicate the experiments in  \cite{Thanapalasingam21}).

\subsection{RQ3: Model Analysis}
\textbf{Hyperbolic Normalization vs. Layer Normalization.} In this experiment, we study the difference between our hyperbolic normalization compared to other existing normalization approaches. For this, we select Layer-norm \cite{ba2016layer} as it is the closest in terms of input/output. Also, in order to study the effect of hyperbolic scaling in our normalization layer, we added constant-norm which is computed as follows: $norm(x) = x / \|x\|$. 
In order to compare the effect of normalization layers, we did not use any extra hyper-parameter rtuning and regularization. We set the embedding size to 64 for both GAT and GCN, and used 4 heads for GAT-models. Table \ref{tab:layer-norm} shows the comparison of different normalization methods on Cora, Pubmed, and Citeseer datasets, and confirms the outperformance of hyperbolic normalization. Since Layer-norm keeps the GAT model in the Euclidean space, we used ADAM optimizer. However, since both hyperbolic normalization and constant normalization have $\|x\|$ as their denominator, we used RiemannianAdam.

\begin{table}[tb!]
    \small
\centering
    \begin{tabular}{l|l|ccc}
    \hline
    \textbf{Base Model} & &\textbf{CR} & \textbf{PM} & \textbf{CS}\\\hline
   \textbf{GCN} & \textbf{Constant} & 67.9 & 76.7 & 68.3\\
         & \textbf{Layer-Norm} & 78.3 & 75.0 & 63.6\\
         & \textbf{Hyp-Norm} & \textbf{80.9} & \textbf{78.0} & \textbf{72.1}\\\hline   \textbf{GAT} & \textbf{Constant} & 75.3 & 73.0 & 61.2\\
         & \textbf{Layer-Norm} & 75.6 & 74.9 & 65.3\\
         & \textbf{Hyp-Norm} & \textbf{76.7} & \textbf{76.3} & \textbf{68.6}\\\hline
    \end{tabular}
    \caption{ Comparison of normalization methods. (Layer-Norm, Hyperbolic Norm, and Constant magnitude). The columns present results on Cora (CR), Pubmed (PM), and Citeseer (CS) datasets using GAT and GCN models.  Note that the provided results are without any hyper-parameter tuning, thus the comparison is only valid column-wise.}
\label{tab:layer-norm}
\end{table}

\vspace{0.05in}
\noindent
\textbf{Optimizer Choice.} In this study, we experimentally validate our observations regarding the choice of optimizers.
Table \ref{tab:optimizer} shows that NGAT works better when RiemannianAdam is used as an optimizer. This shows that, although the parameters are in the Euclidean space, their gradient updates occur in accordance with hyperbolic gradients. Thus, NGAT leverages the architecture of GAT, but behaves more closely to HGAT when it comes to parameter optimization.

\begin{table}[tbp!]
\centering
\resizebox{\linewidth}{!}{
\begin{tabular}{ l|l|ccc}
 \hline
\textbf{Models}&\textbf{Optimizers}&\textbf{Cora}&\textbf{Pubmed}&\textbf{Citeseer}\\\hline
\textbf{GAT}&\textbf{Adam}&\textbf{82.7$\pm$0.2}&78.7$\pm$0.2&\textbf{71.6$\pm$0.3}\\
&\textbf{RAdam}&78.4$\pm$0.3&\textbf{79.0$\pm$0.3}&71.1$\pm$0.3\\\hline
\textbf{HGAT}&\textbf{Adam}&76.8$\pm$0.3&76.0$\pm$0.3&\textbf{68.1$\pm$0.3}\\
&\textbf{RAdam}&\textbf{79.6$\pm$0.3}&\textbf{78.9$\pm$0.3}&\textbf{68.1$\pm$0.3}\\\hline
\textbf{NGAT}&\textbf{Adam}&78.8$\pm$0.3&77.7$\pm$0.3&69.7$\pm$0.3\\
&\textbf{RAdam}&\textbf{83.9$\pm$0.2}&\textbf{79.1$\pm$0.3}&\textbf{73.8$\pm$0.3}\\\hline
\end{tabular}}
\caption{Performance of NGAT, GAT, and HGAT on node classification with RAdam and Adam optimizers.}
\label{tab:optimizer}
\end{table}


\vspace{0.05in}
\noindent
\textbf{Execution Time Comparison.} In this analysis, we demonstrate the impact of hyperbolic normalization in improving the scalability of hyperbolic networks.
We study the computational cost to train our model versus the hyperbolic and Euclidean counterparts.  We ran all the experiments on Quadro RTX 8000 with 48GB memory, and reported the mean of $epoch/second$ for each training experiment for link prediction task in Table \ref{tab:speed-up}. Our model NGCN is 3 times faster than the HGCN on average over the entire training, while NGAT is 5 times faster than its own hyperbolic counterpart (HGAT). NGCN and NGAT are $25\%$ and $12\%$ slower compared to their Euclidean counterparts, respectively. 
We observed the smallest speedup for Pubmed which is due to the large dataset memory footprint. Even in this case, our NGCN model is still more than 2 times faster than HGCN and NGAT is more than 3.5 times faster than HGAT.

\begin{table}[tbp!]
    \small
\centering
    \begin{tabular}{l|l|ccc}
    \hline
    \textbf{Model} & &\textbf{CR} & \textbf{PM} & \textbf{CS}\\\hline
    \textbf{Convolution} & \textbf{GCN} &0.07 & 0.13 & 0.04\\
         & \textbf{HGCN} & 0.29 & 0.33 & 0.21\\
         & \textbf{NGCN} & \textbf{0.08} & \textbf{0.14} &\textbf{0.06}\\\hline
    \textbf{Attention} & \textbf{GAT} & 0.16 & 0.38 & 0.09\\
         & \textbf{HGAT} & 1.10 & 1.34 & 0.81\\
         & \textbf{NGAT} & \textbf{0.19} & \textbf{0.37} & \textbf{0.12}\\\hline
    \end{tabular}
\caption{ Comparison of execution times. The training epoch times (in seconds - lower is better) for Euclidean, hyperbolic, and Pseudo-Poincaré models.}
\label{tab:speed-up}
\end{table}

\begin{figure*}[tb!]
  \centering
  \includegraphics[width=0.75\textwidth]{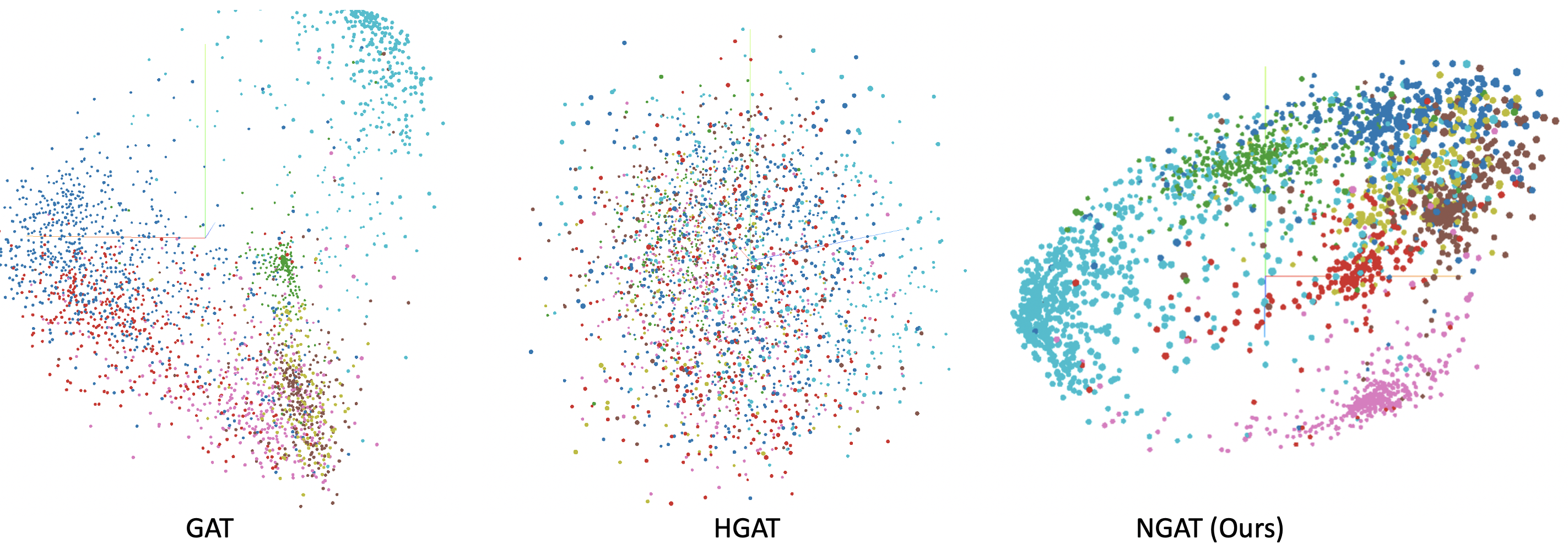}
  \includegraphics[width=0.2\textwidth]{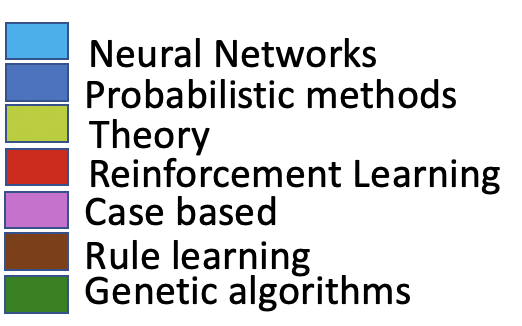}
\caption{
 Embedding visualizations of the Cora citation graph for Euclidean (GAT), hyperbolic (HGAT), and our proposed Pseudo-Poincaré (NGAT) methods. NGAT achieves better separation between the node classes than the corresponding Euclidean and hyperbolic counterparts.}
\label{fig:emb_viz}
\end{figure*}

\subsection{RQ4: Hyperbolic Norm on Deep GNNs}
\cite{li2018deeper} indicates the traditional GNNs such as GCN and GAT achieve their best performance with 2-layer models. Stacking more layers and adding non-linearity tends to degrade the performance of these models. Such a phenomenon is called over-smoothing.  So far, in our experiment, we showed that adding our hyperbolic normalization layer (plus riemannian Adam optimizer) increases the model performance. In order to understand the behavior of our layer on deep GNNs, we experimented on GCNII with 2, 4, 8, 16, 32, and 64 layers.
Please note that since our hyperbolic normalization layer contains tanh, adding the normalization layer to all GCNII causes vanishing gradient descent problem. Thus, we only used our hyperbolic normalization layer in the middle of the model to form and encoder-decoder model (in which the encoder creates a hyperbolic latent space, and a normal decoder interprets the latent space). We picked the same Euclidean hyper-parameters as the original GCNII work \cite{pmlr-v119-chen20v} but did grid search on curvature and scale. Table \ref{tab:gcnii} shows the results for Cora, Citeseer, Pubmed node classification tasks.  As a general observation, the best pick (across models with different layers) of our normalized N-GCNII is either on par or outperforms the best pick of Euclidean GCNII. However, when we look more granular, we observe our model (N-GCNII), outperform the Euclidean counterpart with bugger margin in the shallow models. As the number of layers increases, the performance gap reduces as well. We believe this is due to the fact that deeper model are better at learning smooth manifolds which is consistent with studies on NLP indicating the latent space of deep models shows a degree of hyperbolicity \cite{chen2021probing}. 
Moreover, we observed that as the model gets deeper, applying Riemannian Adam is less effective (for models deeper than 8 layers, ADAM performs better even on our models).

\begin{table}[tb!]
\centering
\resizebox{\linewidth}{!}{
\begin{tabular}{ ll|cccccc} \hline\textbf{}&\small{\textbf{Models}}&\small{\textbf{2}}&\small{\textbf{4}}&\small{\textbf{8}}&\small{\textbf{16}}&\small{\textbf{32}}&\small{\textbf{64}}\\\hline
\small{\textbf{Cora}}
&\small{\textbf{GCNII}}& 83.1 & 83.6 & 84.6 & 85.6 & \small{\textbf{86.4}} & 86.1 \\
&{\textbf{NGCNII}}& 83.6 & 84.5 & 85.4 & 85.8 & \small{\textbf{86.3}} & 86.1 \\\hline
\small{\textbf{Citeseer}}
&\small{\textbf{GCNII}}& 68.7 & 69.1 & 71.3 & 73.4 & 73 & 73.2 \\
&\small{\textbf{NGCNII}}& 69.8 & 70.2 & 68.8 & 73.6 & \small{\textbf{74.1}} & 73.1 \\\hline
\small{\textbf{Pubmed}}
&\small{\textbf{GCNII}}& 78.6 & 78.5 & 79 & \small{\textbf{80.8}} & 80.2 & 80.4 \\
&\small{\textbf{NGCNII}}&79.4 & 79.2 & 80 & 80.5 & \small{\textbf{80.9}} &80.5 \\\hline
\end{tabular}
}
\caption{ \label{tab:gcnii} The behavior of hyperbolic normalization on deep GNNs.}
\end{table}

\subsection{RQ5: Embedding Visualizations}
We visually demonstrate that our approach of learning in Euclidean space, indeed preserves the hierarchical characteristics of hyperbolic space. Towards this goal, we present the visualization of node embeddings from the Cora citation graph. We choose Cora, since it is a citation-based graph and does not contain explicit hierarchical (is-a) relations. We contrast the embeddings learned by GAT (a method based on Euclidean space), with  HGAT (learning in hyperbolic space) with the proposed NGAT approach. Figure \ref{fig:emb_viz} shows that NGAT ensures clearer separation between node classes in comparison to its Euclidean and hyperbolic counterparts.

\section{Conclusion}
\label{sec:conclusion}
In this paper, we showed that a Euclidean model can be converted to behave as a hyperbolic model by simply (1) connecting it to the hyperbolic normalization layer and (2) using a Riemannian optimizer. 
From the training perspective, this approach allows the trainable parameters to interact in the Euclidean vector space, consequently enabling the use of Euclidean regularization techniques for the model training. Furthermore, the training of our proposed approach is significantly faster compared to the existing hyperbolic methods. This allows the hyperbolic networks to be more flexible in practice to a wider variety of graph problems.
From task perspective, we removed the implicit assumption of having spatial locality of the adjacent nodes (in all dimensions), to be a prerequisite of the hyperbolic model training. This allows the framework to be applied on a wider range of tasks.

\bibliographystyle{named}
\bibliography{ijcai23}


\appendix

\section*{Appendices}

\section{Proof of Lemma 2}
\label{app:proof-lemma-2}
Given $x \in \mathbb{R}^n$ is a Euclidean equivalent of Poincaré ball's point $p \in \mathbb{B}^n_c$ in the tangent space at origin $\mathcal{T}_{0^n}\mathbb{B}_c^n$. The relation between $x$ and $p$, as defined by \cite{ganea2018hyperbolic} as;
\begin{align}
    \nonumber p = exp_0^c(x) &= \tanh(\sqrt{c}\|x\|)\frac{x}{\sqrt{c}\|x\|},
\end{align}
\begin{align}
    \nonumber x = log_0^c(p) = \tanh^{-1}(\sqrt{c}\|p\|)\frac{p}{\sqrt{c}\|p\|},
\end{align}
Simplifying with $\omega(x) = \frac{\tanh(\sqrt{c}\|x\|)}{\sqrt{c}\|x\|}$ and $\omega_{\mathbb{B}_c^n}(p) = \frac{\tanh^{-1}(\sqrt{c}\|p\|)}{\sqrt{c}\|p\|}$ results in:
\begin{align}
    &p = \omega(x)x \text{ and } x = \omega_{\mathbb{B}_c^n}(p)p
\end{align}

\section{Proof of Lemma 3 and Theorem 1}
\label{app:proof-lemma-3-theorem-1}
Let us first look at the case of n = 1, i.e., a single hyperbolic feed-forward layer. This is defined by \cite{ganea2018hyperbolic} as;
\begin{equation}
    F_1^{\otimes_c}(p) = f_1^{\otimes_c}(p) = exp_0^c(f_1(log_0^c(p)))
\end{equation}
Reformulating this with Lemma 1,
\begin{align}
    \nonumber F_1^{\otimes_c}(p) &= \omega(f_1(log_0^c(p)))f_1(log_0^c(p));
\end{align}
\begin{align}
    \text{Substituting, }x = log_0^c(p);~~F_1^{\otimes_c}(p) &= \omega(f_1(x))f_1(x);
\end{align}
Extending this to $n=2$,
\begin{align}
    \nonumber F_2^{\otimes_c}(p) &= f_2^{\otimes_c}(f_1^{\otimes_c}(x))
    = f_2^{\otimes_c}(\omega(f_1(x))f_1(x))\\
    \nonumber&\text{Given that $f_1(x)$ is a linear function; }\\
    f(ax) &= af(x);\\
    \nonumber F_2^{\otimes_c}(p) &= \omega(f_1(x))f_2^{\otimes_c}(f_1(x));\\
    \nonumber F_2^{\otimes_c}(p) &= \omega(f_1(x))\omega(f_2(x))f_2(f_1(x));\\
    \nonumber F_2^{\otimes_c}(p) &= \omega(f_2(x))\omega(f_1(x))F_2(x);\\
    \nonumber &\text{Substituting; }\\
    \Omega(F_2(x)) &= \omega(f_1(x))\omega(f_2(x));\\
    F_2^{\otimes_c}(p) &= \Omega(F_2(x))F_2(x);
\end{align}
Extending the above formulation to arbitrary $n$,
\begin{align}
    \nonumber F_n^{\otimes_c}(p) &= \omega(f_n(x))...\omega(f_2(x))\omega(f_1(x))F_n(x);
\end{align}
Reformulating this with Lemma 1, we get the conclusion of Lemma 2;
\begin{align}
    F_n^{\otimes_c}(p) &= \exp_0^c(F_n(log_0^c(p)));
\end{align}
\begin{align}
    \nonumber &\text{Substituting; }\\
    \Omega(F_n(x)) &= \omega(f_n(x))...\omega(f_2(x))\omega(f_1(x));\\
    \nonumber &\text{We get the conclusion of Theorem 1;}\\
    \nonumber F_n^{\otimes_c}(p) &= \Omega(F_n(x))F_n(x);
\end{align}

\end{document}